\documentclass[letterpaper, 10 pt, conference]{ieeeconf}  
\IEEEoverridecommandlockouts                              

\IEEEoverridecommandlockouts    

\usepackage[pdftex]{graphicx}
\graphicspath{{figures/}}
\usepackage{hyperref}
\hypersetup{
    colorlinks=true,
    linkcolor=black,
    citecolor=black,
    filecolor=black,
    urlcolor=black,
}
\usepackage[T1]{fontenc}
\usepackage[utf8]{inputenc}
\usepackage{csquotes}
\usepackage[english]{babel}
\usepackage[export]{adjustbox}
\usepackage{caption}
\usepackage{subcaption}
\usepackage[colorinlistoftodos]{todonotes}
\usepackage{lipsum}
 \usepackage{amsmath} 

\usepackage{mathtools}
 \usepackage{amssymb}
\usepackage{placeins}
\usepackage{pgfplots}
\pgfplotsset{compat=1.11}
\usepackage{algorithm}
\usepackage{algorithmic}
\usepackage{array,booktabs}
\usepackage{tikz}
\usepackage{multirow,graphicx}
\usepackage{makecell}
\usetikzlibrary{matrix,decorations.pathreplacing}
\usepackage[style=ieee,
            doi=false,
            url=false,
            mincitenames=1,
            maxcitenames=1,
            minbibnames=6,
            maxbibnames=6,
            backend=bibtex]{biblatex}  
\renewcommand{\S}{\mathcal{S}}
\newcommand{\A}{\mathcal{A}}
\renewcommand{\O}{\mathcal{O}}

\newcommand{\tm}{\ensuremath{P}}
\newcommand{\belo}{\ensuremath{\mathcal{T}}}
\newcommand{\om}{\ensuremath{\mathcal{Z}}}

\renewcommand{\rm}{\ensuremath{R}}
\newcommand\distEq{\stackrel{\mathclap{{D}}}{=}}
\newcommand{\mpsc}{\frac{\text{m}}{\text{s}^3}}
\newcommand{\mpsq}{\frac{\text{m}}{\text{s}^2}}
\newcommand{\mps}{\frac{\text{m}}{\text{s}}}

\title{\LARGE \bf
	Minimizing Safety Interference for Safe and Comfortable Automated Driving with Distributional Reinforcement Learning
}

\author{Danial Kamran$^{1}$, Tizian Engelgeh$^{2}$, Marvin Busch$^{2}$, Johannes Fischer$^{1}$ and Christoph Stiller$^{1}$
	\thanks{$^{1}$Authors are with Institute of Measurement and Control Systems, Karlsruhe Institute of Technology (KIT), 76131 Karlsruhe, Germany, {\tt\small \{danial.kamran, johannes.fischer, stiller\}@kit.edu }}%
\thanks{$^{2}$Authors are students at  Karlsruhe Institute of Technology (KIT), 76131 Karlsruhe, Germany, {\tt\small \{tizian.engelgeh,marvin.busch\}@student.kit.edu}}%
}

\begin{document}

\maketitle

 \pubid{\begin{minipage}{\textwidth}~\\[12pt] \centering%
   \copyright~2021 IEEE. Personal use of this material is permitted. Permission from IEEE must be obtained for all other uses, including reprinting/republishing this material for advertising or promotional purposes, collecting new collected works for resale or redistribution to servers or lists, or reuse of any copyrighted component of this work in other works.
 \end{minipage}}
 \pubidadjcol

\pagestyle{empty}

\begin{abstract}
Despite recent advances in reinforcement learning (RL), its application in safety critical domains like autonomous vehicles is still challenging. 
Although punishing RL agents for risky situations can help to learn safe policies, it may also lead to highly conservative behavior. 
In this paper, we propose a distributional RL framework in order to learn adaptive policies that can tune their level of conservativity at run-time based on the desired comfort and utility. 
Using a proactive safety verification approach, the proposed framework can guarantee that actions generated from RL are fail-safe according to the worst-case assumptions. 
Concurrently, the policy is encouraged to minimize safety interference and generate more comfortable behavior. 
We trained and evaluated the proposed approach and baseline policies using a high level simulator with a variety of randomized scenarios including several corner cases which rarely happen in reality but are very crucial. 
In light of our experiments, the behavior of policies learned using distributional RL can be adaptive at run-time and robust to the environment uncertainty.
Quantitatively, the learned distributional RL agent drives in average 8 seconds faster than the normal DQN policy and requires 83\% less safety interference compared to the rule-based policy with slightly increasing the average crossing time.
We also study sensitivity of the learned policy in environments with higher perception noise  and show that our algorithm learns policies that can still drive reliable when the perception noise is two times higher than the training configuration for automated merging and crossing at occluded intersections.


\end{abstract}

\section{Introduction}
\label{introduction}
Reinforcement learning (RL) has gained more attention recently in order to solve complex decision making problems in robotics.
Specifically for self-driving vehicles, long term optimal policies are learned using this approach for multiple scenarios such as  lane changing or merging in highways \cite{Mirchevska_lane_change, safe_multi_agent,tang2020worst} or yielding at unsignalized intersections \cite{isele_navigating,learning_negotiating,isele_safe_rl,bouton_safe_rl,kamran2020risk}.
Benefitting from the high representational power provided by neural networks to learn the future return of each action, learning-based policies can provide optimal behavior which is more generic and scalable compared to POMDP-based approaches like \cite{hubmann2019pomdp} and also more intelligent than rule-based approaches \cite{ttc,bbf}.

Safety is one of the most important challenges for learning-based policies in critical applications like automated driving. 
Although the learned policies are prevented to generate risky behavior during training by receiving big punishments for having collisions \cite{isele_navigating,learning_negotiating} or being in risky situations \cite{isele_safe_rl,kamran2020risk}, there is no guarantee that the learned policy is always safe after training.
Another important challenge for the learning-based policies is to apply them in environments which are more challenging than the training environment due to higher sensor noise in perception or more severe sensor occlusions. 
Although one can try to learn a policy which reduces the amount of risk to be still safe in more challenging scenarios like \cite{kamran2020risk}, such a policy often is too conservative when the uncertainty is high.
This dilemma between more conservative but slower policies and risky but faster policies is nicely elaborated by Isele et. al in \cite{isele_safe_rl} where higher timeout punishments as part of the reward function resulted in a faster but more risky policy.
The origin of this problem lies in the difficulty of shaping the reward to guarantee having safe policies in gradient based learning approaches as discussed in \cite{safe_multi_agent}. 

\begin{figure}[t]
	\includegraphics[width=\linewidth]{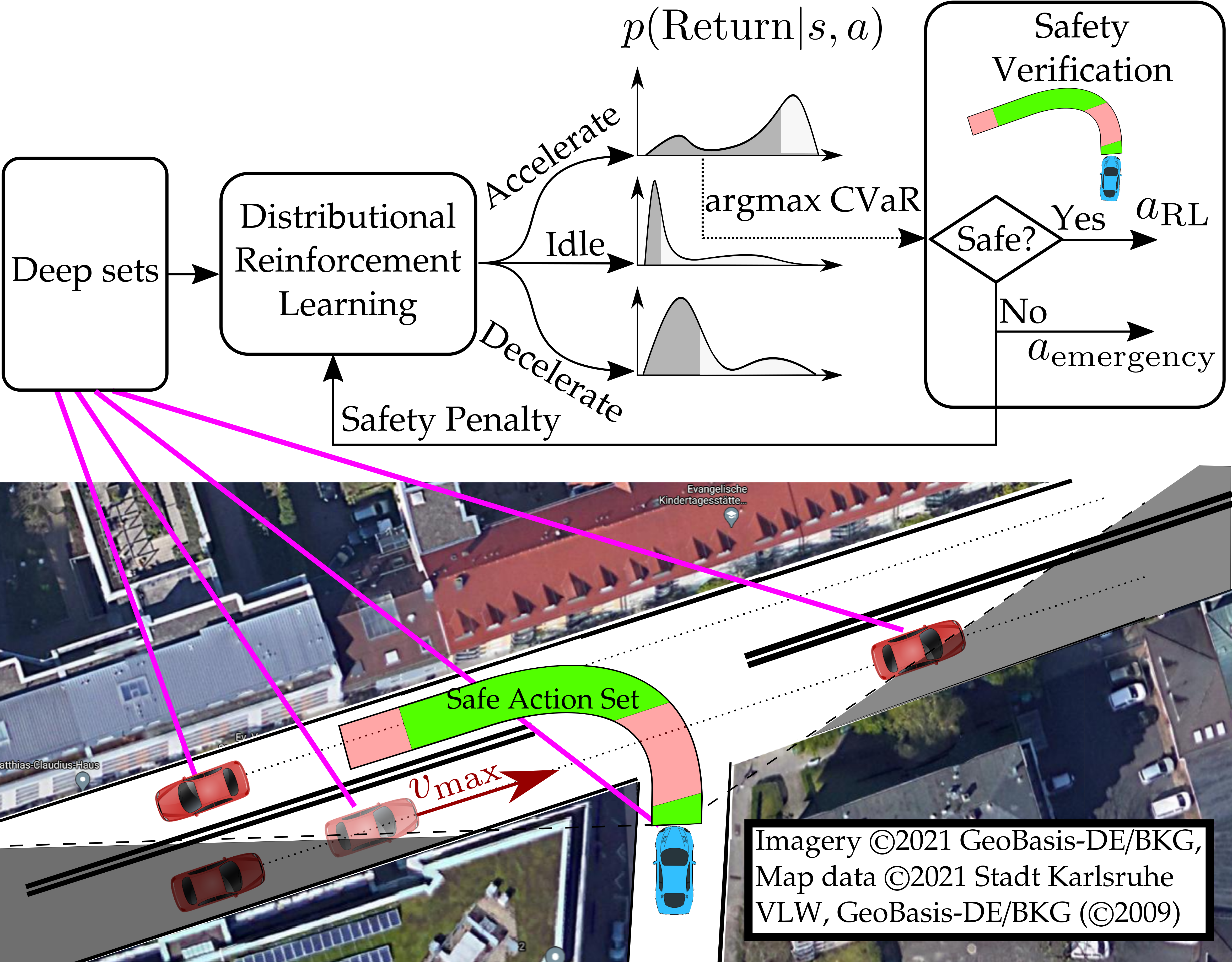}
	\caption{Overview of the merging scenario and the proposed  distributional reinforcement learning framework for automated navigation at similar occluded intersections (background image source: \cite{google_map}).}
	\label{fig:scenario}
\end{figure}

\pubidadjcol

Another approach to address safety for RL policies is to utilize a safety layer which filters out unsafe actions as proposed in \cite{Mirchevska_lane_change,alshiekh2017safe, lazarus_runtime_safe_rl}. 
This safety layer must be easily verifiable without any black boxes such as deep neural networks inside its architecture which are hard to verify \cite{liu_algorithms_dnn_verification}.
In \cite{Mirchevska_lane_change} authors used safety constraints in order to prevent unsafe lane changes from a RL agent. As the fail-safe action, the automated vehicle will keep the lane whenever lane change actions are unsafe and in order to prevent unnecessary lane changes, the RL agent will receive punishments for every lane change decision.
In \cite{alshiekh2017safe}, the authors showed that safety verification can also help the RL agent to converge faster in addition to guaranteeing safety. They studied the effect of applying the safety layer before or after the RL agent as \textit{preemptive} and \textit{Post-Posed Shielding} mechanisms and also specifying penalties for every safety intervention on the performance of the learned policy.

In the context of automated driving, uncertainties due to perception noise or ambiguous behavior of other participants need to be considered inside the safety verification constraints.
However, there is a trade-off between the level of safety that is guaranteed and efficiency \cite{sha_simplicity_control}.
Assuming that always the worst case perception noise exists and other drivers are all distracted will result in unnecessary safety interruptions which actually prevent the RL agent to utilize its learning skills.
On the other hand, one can increase the capability of intervention (e.g. deceleration with -10 \text{$m/s^2$}) and only prevent unsafe actions just before a hazardous event is going to happen which will result in rare interventions but with uncomfortable feelings for the passengers.



Our contribution is to address safety, scalability and comfort as the main intertwined challenges for automated driving under uncertainty. 
We introduce safe distributional RL which considers maximum uncertainty and capability inside  safety verification layer.
During training the RL agent is punished for every safety intervention similar to \cite{lazarus_runtime_safe_rl, alshiekh2017safe}. 
However, the main difference is it learns distributions instead of expected values for each action return which helps to provide risk-aware policies that are able to adapt their conservativity according to the existing uncertainty in the environment. 
This feature makes the learned policy generic and applicable for  high uncertainty levels that can rarely happen in realistic applications, for example when the sensor range is severely occluded due to a parked vehicle or it has high uncertainty.

After explaining preliminaries of our work in section \ref{sec:preliminaries}, we describe the proposed worst-case based safety verification layer in section \ref{sec:proactive_safety} and the safe distributional RL framework in section \ref{sec:safe_RL}. Finally, efficiency of the proposed method is compared with a regular RL and a rule-based agent as baseline policies in section \ref{sec:results_and_evaluation}.

\section{Preliminaries}\label{sec:preliminaries}
\subsection{Reinforcement Learning}

Reinforcement learning problems are typically formulated as a
Markov Decision Process (MDP) 
$(\S, \A, \tm, \rm, \gamma)$ \cite{howard_mdp}.
In this framework, the environment is described by a state
$s_t\in\S$ 
and the agent chooses an action
$a_t\in\A$ 
at each discrete time step $t$.
The action selection process is described by a policy $\pi$ which is a mapping from states to actions
(or to distributions over actions).
The distribution of the successor state $s_{t+1}$ is defined by the transition model
$s_{t+1}\sim\tm(s_t,a_t)$ based on the current state and chosen action.
In every step the agent receives a reward 
$r_t=\rm(s_t,a_t)$ and
its goal is to maximize the cumulative discounted reward (also called return)
$\sum_{t=0}^{\infty} \gamma^t r_t$,
where future rewards are discounted by the factor
$\gamma \in [0,1)$.

In many real-world applications the agent does not have full knowledge of the environment's state 
but instead receives only noisy observations of some state variables.
Such situations can be described by a
Partially Observable Markov Decision Process (POMDP)
$(\S, \A, \O, \tm, \om, \rm, \gamma)$, 
where the agent receives an observation
$o_t \in \O$
at each time step.
The observation depends on the current state and chosen action and is distributed according to the observation model
$o_t \sim \om(s_t, a_t)$.
The agent's goal is still to maximize the return but with the additional challenge of an unknown environment state.
Since each observations contains only limited information on the true environment state, policies for a POMDP therefore also depend on previous actions and observations.

A famous approach to find the optimal RL policy is $Q$-learning
\cite{qlearning} which tries to maximize the expected future return defined as:
\begin{equation*} 
	\label{eq:q}
	Q^\pi(s_t,a_t) = \mathbb{E}_{s_i\sim\tm}\left[ \rm(s_t,a_t) + \sum_{k=1}^\infty \gamma^{k} \rm(s_{t+k}, \pi(s_{t+k})) \right], 
\end{equation*}
by choosing action $a_t$ in state $s_t$ and following policy $\pi$ for the next states \cite{sutton_barto_rl}.
Using the Bellman equation \cite{bellman_dynamic}, the optimal value function can be represented as:
\begin{equation*}
	Q^*(s_t,a_t) = \mathbb{E}_{s_i\sim\tm}\left[ \rm(s_t,a_t) + \gamma \max_{a^\prime} Q^*(s_{t+1}, a^\prime)) \right].
\end{equation*}

In Deep Q networks (DQN) \cite{dqn}, deep neural networks are utilized to learn the Q values for each state and action over samples from a replay buffer.
We use DQN as one of baselines in our experiments.



\subsection{Distributional Reinforcement Learning}
Distributional Reinforcement Learning \cite{bellemare2017distributional_RL} tries to learn return distribution $Z^\pi(s_t,a_t) \distEq R(s_t,a_t) + \sum_{k=1}^\infty \gamma^{k} \rm(s_{t+k}, \pi(s_{t+k}))$ instead of its expected value. 
Note that here $\distEq$ indicates equality in distribution.
The value distribution can be computed using dynamic programming based on the \textit{distributional Bellmann equation} \cite{bellemare2017distributional_RL} :
\begin{equation}
Z^\pi(s,a) \distEq R(s,a) + \gamma Z^\pi(S^\prime,A^\prime),
\end{equation}
where $S^\prime$ and $A^\prime$ are distributed according to $\tm(.|s, a)$ and $\pi(.|s^\prime)$, respectively.
This equation is shown in \cite{bellemare2017distributional_RL} to be a contraction in the Wasserstein metric.

Similar to the regular RL, the \textit{Distributional Bellman optimality equation} is applied:
\begin{equation*}
\belo Z(s,a) \distEq R(s,a) + \gamma Z(S^\prime,\arg\max_{a^\prime\in\A}\mathbb{E}Z(S^\prime,a^\prime)),
\end{equation*}
with $S^\prime$  distributed according to $\tm(\,\cdot\,|s, a)$.
In order to learn the optimal return distribution, Bellemare et al. parameterized it as a categorical distributions over a fixed set of equidistant points \cite{bellemare2017distributional_RL}.
Their approach, called C51, minimizes the Kullback-Leibler divergence to the distributional Bellman targets, which, however, was not a contraction in the Wasserstein metric. 
Later Dabney et al. proposed to learn return distributions through \textit{Quantile Regression} (QR-DQN) on a fixed set of quantiles, minimizing the Wasserstein distance to the distributional Bellman targets \cite{dabney2018quantile}.
In a newer approach,  Dabney et al. introduced Implicit Quantile Networks (IQN) \cite{dabney2018implicit} to approximate the quantile function $F^{-1}_Z(\tau)$ for the random
variable $Z$. 
Assuming $\tau \sim U(\left[0, 1\right])$, the return distribution can then be sampled from $F^{-1}_Z(\tau)$ as samples from implicitly defined return distribution.
The main advantage of IQN is that any distortion risk measure $\beta:\left[0, 1\right] \to \left[0, 1\right]$ can be incorporated to compute distorted expectation of $Z$:
\begin{equation*}
	Q_\beta(s_t, a_t)= \mathbb{E}_{\tau \sim U(\left[0, 1\right])} \left[ Z_{\beta(\tau)}(s_t, a_t)\right],
\end{equation*}
and the risk-sensitive greedy policy: 
\begin{equation*}
\pi_\beta(s_t) = \arg\max_{a\in\A}Q_\beta(s_t, a).
\end{equation*}
In this paper we utilize IQN implementation of distributional RL since it allows to explore risk-sensitive policies $\pi_\beta$ during training which helps us to learn a family of risk-averse policies using only one neural network.

\begin{figure}[t]\centering
	\fontsize{8pt}{11pt}\selectfont
	\def\svgwidth{0.99\linewidth}
	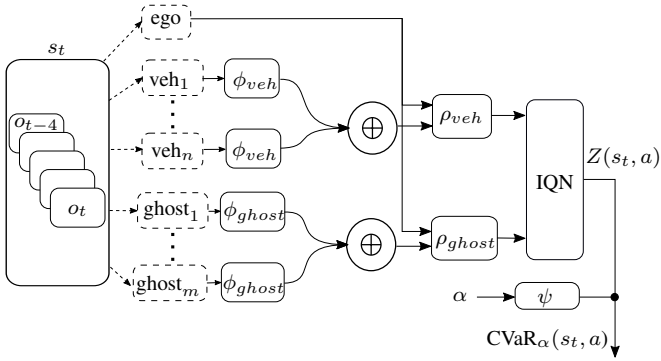
	\caption{Proposed Deep sets architecture for extracting permutation invariant and scalable features for the distributional RL network.}\label{fig:architectures}
\end{figure}

\subsection{Observation and State Model}
As proposed in \cite{kamran2020risk}, we can represent the whole situation at an occluded intersection using this matrix:

\begin{equation}
o_t = \begin{tikzpicture}[baseline, decoration=brace]
\matrix (m) [matrix of math nodes,left delimiter=[,right delimiter={]^T}] {
	d_{\textrm{e},\textrm{stl}} & d_{1} & ... & d_{n} & d_{o_1} & ... & d_{o_m}  \\
	v_{\textrm{e}} & v_1 &  ... & v_n & v_{o_1} & ... & v_{o_m}  \\
	d_{\textrm{e},\textrm{goal}} & d_{\textrm{e},1} & ... & d_{\textrm{e},n} & d_{\textrm{e},o_1} & ... & d_{\textrm{e},o_m} \\
};
\draw[decorate,transform canvas={yshift=0.5em},thick] (m-1-1.north west) -- node[above=2pt] {$ego$} (m-1-1.north east);
\draw[decorate,transform canvas={yshift=0.5em},thick] (m-1-2.north west) -- node[above=2pt] {$vehicles$} (m-1-4.north east);
\draw[decorate,transform canvas={yshift=0.5em},thick] (m-1-5.north west) -- node[above=2pt] {$lanes$} (m-1-7.north east);
\end{tikzpicture}
\label{eq:observation}
\end{equation}
where $v_{\textrm{e}}$ is the ego vehicle velocity and $d_{\textrm{e},\textrm{stl}}$, $d_{\textrm{e},\textrm{goal}}$ are its distance to the stop line and to the other side of the intersection.
$d_i, v_i$ are distance and velocity of every observable vehicle to their conflict zones and $d_{o_i}, v_{o_i}$ are information for ghost vehicles indicating maximum visible distance and allowed velocity on every intersecting lane.
Finally, $d_{\textrm{e},i}$ ($d_{\textrm{e},o_i}$) represents the distance of the ego vehicle to every (ghost) vehicle. 
We extend this representation for merging scenarios where we also observe the distance and velocity of the closest front vehicle on the ego vehicle lane.

Finally, we provide a $k$-Markov approximation \cite{bouton2018k_markov} of the POMDP as input to the RL agent in order to enable $Q$-learning similar to \cite{kamran2020risk}:
\begin{equation}
s_t = 
\begin{bmatrix}
o_t & o_{t-1} & ... & o_{t- (k-1)}
\end{bmatrix}
\end{equation}
In this way the neural network and its training process become less complex compared to other implementations for POMDPs like \cite{learning_negotiating} which utilize Long Short-Term Memory (LSTM) cells \cite{lstm} to incorporate past information.
In our experiments we supplied the last $k=5$ observations to the network.
Note that the direction of vehicles is specified by their velocity sign and their intention can be estimated from states history.
Other traffic participants can also be added into this model like pedestrians or cyclist in order to provide more generic model.
This representation is more efficient compared to the grid-based representations used in \cite{isele_navigating, kamran2019learning} which require deep convolutional layers to extract useful features which are sensitive to the roads topology and irrelevant traffic participants.

\subsection{Scalable Reinforcement Learning}
The dimension of the state representation can exponentially grow for complex scenarios when the number of vehicles $n$ or intersecting lanes $m$ increase.
Another challenge here is that permutation of input elements can change which may cause the network to have different reactions for the same scenario with different input permutations \cite{huegle2019dynamic}.

In order to address this problem, we use Deep sets  \cite{zaheer2017deepsets} architectures that decouple the network size of machine learning algorithms from the number of input elements. 
Deep sets approach has already been applied to learn a lane change policy with DQN for highway scenarios in \cite{huegle2019dynamic}. 
In this paper, we propose a Deep sets architecture for automated navigation at occluded merging and crossing scenarios (Figure \ref{fig:architectures}).
A representation is calculated for each type of the input element (real vehicle or ghost vehicles) with the $\phi_{real}$ and $\phi_{ghost}$ networks. 
After that, all features for each element type are combined with a permutation invariant operator, which we used sum of them. 
Finally, $\rho_{real}$ and $\rho_{ghost}$ networks extract fixed size features from the combination of each type of the input element with  the ego vehicle state. 

\section{Proactive Safety Verification}\label{sec:proactive_safety}
In order to verify safety, we propose a proactive safety verification approach which provides safe actions for every state $s\in \S$.
We call it proactive while the worst-case scenario is always considered to  prevent being in an unsafe state in the future.
\subsection{Worst-Case Scenarios for Proactive Safety Verification} 
We define a set of worst-case scenarios  $\mathcal{H}$ that can happen during automated driving at intersections:
\begin{enumerate}
	\item On the intersecting lane $l_i$, an occluded vehicle is driving with velocity $v_{\text{max}}$ at the closest occluded distance to the conflict zone.
	\item On the intersecting lane $l_i$, an observed vehicle has an estimation error of $3\sigma_\text{d}$ and $3\sigma_\text{v}$ for its distance and velocity and accelerates with $a_{\text{max}}$  to reach velocity of $v_{\text{max}}$.
	\item On the ego lane $l_{ego}$, an observed vehicle in front of the ego vehicle has an estimation error of $3\sigma_\text{d}$ and  $3\sigma_\text{v}$ for its distance and velocity and  decelerates with $a_{\text{min}}$  to reach zero velocity.
\end{enumerate}
By assumptions in $\mathcal{H}$ we over-approximate the state of an occluded vehicle at intersecting lanes similar to \cite{pio_2018_occlusion} and for detected vehicles as well.
In addition to \cite{pio_2018_occlusion}, we over-approximate vehicles distance and velocity estimation error as $\mathcal{N}_a^t(\mu,\sigma_\text{d}^{2})$ and $\mathcal{N}_a^t(\mu,\sigma_\text{v}^{2})$ respectively which are denoted as truncated Gaussian distribution with support
$[\mu - a\sigma, \mu + a\sigma]$.
Therefore, the distance and velocity measurement errors are modeled with truncated Gaussian distributions 
$\mathcal{N}_3^t(d,\sigma_\text{d}^{2})$ and $\mathcal{N}_3^t(v,\sigma_\text{v}^{2})$, respectively,
i.e.\ with a $3\sigma$ range around the true value.

\subsection{Feasibility of Emergency Maneuvers for Safety Verification} 
For every intersecting lane $l_i$ and also the ego vehicle lane $l_{ego}$ inside the current state $s$, feasibility of executing one of the following two emergency maneuvers is evaluated to verify the state $s$ as a proactive safe state (PSS) according to the worst-case assumptions  $\mathcal{H}$:
\begin{itemize}
	\item Emergency Stop  Maneuver: Ego vehicle is able to stop at a distance $d_\mathrm{FS}$ bigger than $SG_d$ before the conflict zone. 
	\item Emergency Leave Maneuver: Ego vehicle is able to leave the intersection with a time headway $t_\mathrm{HW}$ bigger than $SG_t$.
\end{itemize}
Thus, PSS verification can be formulated as:
\begin{equation}\label{eq:PSS}
\text{PSS}(s)=
\left(\bigwedge\limits_{l_i \in s} d_\mathrm{FS}>SG_d \right)  \bigvee 
\left(\bigwedge\limits_{l_i \in s} t_\mathrm{HW}>SG_t \right),
\end{equation}
with $SG_d$ and $SG_t$ the minimum safety gap required for stop and leave maneuvers.
%
Note that here we validate safety according to all worst case assumptions in $\mathcal{H}$, i.e. if only one of them or all of them happens the situation should remain safe.
We set $SG_d=SG_t=0.5$ in our experiments which resulted in collision free maneuvers during training and evaluation of all policies.

\subsection{Proactive Safety Verification of Actions in POMDP}
We assume an agent selects action $a\in\A$ for an observation $o\in\O$ from our POMDP model.
In order to verify safety of $a$, we need to make sure that  it results in a proactive safe state even if the worst-case scenario happens.
Therefore, we simulate the future state of the ego vehicle by executing $a$ and the state of other vehicles with the worst-case assumptions in $\mathcal{H}$ until the next decision time.
We call this simulated state as $\hat{s}_{\mathcal{H},o}^a$.
If $\hat{s}_{\mathcal{H},o}^a$ is a PSS, then action $a\in\A$ is verified as a proactive safe action (PSA):
\begin{equation}\label{eq:PSA}
\text{PSA}(s, a)=
\begin{cases}
\text{True}, & \text{if}\ \text{PSS}(\hat{s}_{\mathcal{H},o}^a), \\
\text{False}, & \text{otherwise}.
\end{cases}	
\end{equation}

The main advantage of the proposed worst-case proactive safety verification compared to reachability-based \cite{althoff2014online} and prediction-based approaches \cite{isele_safe_rl} for safety is that it only needs to predict the whole situation for one time step and only for the worst-case scenarios defined in $\mathcal{H}$ instead of all possible maneuvers for all participants in the whole future horizon. 
Since for each intersecting lane, the proposed safety verification approach only considers the closest vehicle to the conflict zone, it has computation complexity of $O(m+1)$ where $m$ is the number of intersecting lanes in the scenario.
This complexity is much smaller than the safety verification strategy proposed in \cite{isele_safe_rl} which has complexity of $O(nT)$ where $n>>m$ is the number of agents and $T$ is the time horizon considered for safety verification in that approach.

\section{Learning Safe and Comfortable Policies for Automated Driving Under Uncertainty}\label{sec:safe_RL}

Similar to \cite{safe_multi_agent}, we propose to apply safety as a hard constraint in the decision making problem trying to minimize other costs such as utility  or comfort as soft constraints.
For that, we can use the PSS concept in order to find all safe actions among the set of available actions in $\A$ as $\A_\text{PSA}$:
\begin{equation}
\A_\text{PSA}(s) = \{a\in\A | \text{PSA}(s,a)=\text{True}\},
\end{equation} 
and search for the safe action with the lowest cost regarding comfort and utility.
Here we consider actions as jerk commands applied to the automated vehicle similar to \cite{werling2010optimal, muller2019risk} which help to provide comfortable maneuvers by moderating jerk.
Therefore, we set possible discrete actions as
$\A=\{-1.5\frac{m}{s^3}, 0.0\frac{m}{s^3}, 1.5\frac{m}{s^3} \}$.

\subsection{Safe Rule-based Policy}
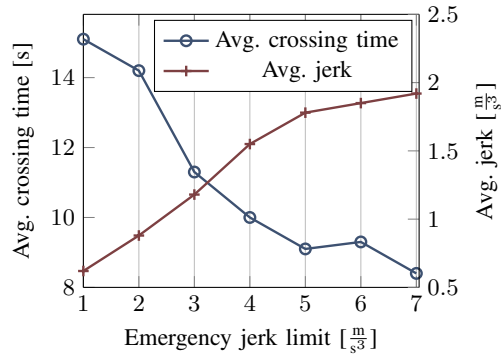
\begin{figure}[t]
	\centering
	\fontsize{9pt}{11pt}\selectfont
	\begin{tikzpicture}
	\begin{axis}[xlabel=Emergency jerk limit \text{[$\frac{\text{m}}{\text{s}^3}$]}, xmajorgrids=true,ymin=8,ymax=,xmax=7.05,xmin=1,
	xtick={1,...,7},
	ylabel=Avg. crossing time \text{[s]} ,	width=0.70\linewidth,
	axis y line*=left]
	\addplot[mark=o,color={rgb:red,88;green,117;blue,164},line width=0.9pt] coordinates {
		(1.0, 15.1)
		(2.0, 14.2)	
		(3, 11.3)				
		(4, 10.)
		(5, 9.1)				
		(6, 9.3)				
		(7, 8.4)						
	};\label{plot_one}

	\end{axis}
	
	\begin{axis}[xlabel=Emergency jerk limit,ymin=0.5,ymax=2.5,xmax=7.05,xmin=1,ylabel=Avg. jerk \text{[$\frac{\text{m}}{\text{s}^3}$]} ,	width=0.70\linewidth,
	axis y line*=right, axis x line=none]
	\addlegendimage{/pgfplots/refstyle=plot_one}\addlegendentry{Avg. crossing time}
	
	\addplot[mark=+,color={rgb:red,181;green,93;blue,96},line width=0.9pt] coordinates {
		(1.0, 0.62)
		(2.0, 0.88)	
		(3, 1.18)				
		(4, 1.55)
		(5, 1.78)				
		(6, 1.85)				
		(7, 1.92)						
	};
	\addlegendentry{Avg. jerk}
	\end{axis}
	
	\end{tikzpicture}
	\caption{Effect of increasing the jerk limit on the average crossing time and average jerk for the proactive safe policy.}\label{fig:time_jerk}
\end{figure}

We implemented a rule-based policy which selects the fastest safe action from $\A$ and in case of emergency selects an emergency action with the lowest jerk.
The maximum emergency jerk limit can be increased which result in higher utility in cost of less comfortable drivings due to higher average jerk (Figure \ref{fig:time_jerk}). 
On the other hand, when emergency jerk limit is reduced, the policy becomes slower, more comfortable and more conservative, since emergency maneuvers are limited.
  
\subsection{Minimizing Safety Interference with Reinforcement Learning}
RL, thanks to its ability to provide long-term optimal actions, can trade-off between utility and comfort based on the preferences defined in its reward function:
\begin{equation}
\rm(o, a)=
\begin{cases}
$ 1$, & \text{if reach goal} \\
-\lambda, & \text{if}\ a \not\in {A}_{\text{PSA}}(o) \\
$ 0$, & \text{otherwise}
\end{cases}	
\label{eq:reward}
\end{equation}
By such reward function, safety interventions are discouraged preventing situations that an uncomfortable emergency action with higher jerk instead of the unsafe RL action needs to be executed. 
The proposed framework has two main differences compared to other safe RL frameworks like \cite{alshiekh2017safe, Mirchevska_lane_change}.
First, the emergency action is not necessarily a member of $A_\text{RL}$, therefore the training episode is terminated when a safety interference is required and the RL unsafe action will not be replaced in the replay buffer. 
Secondly, the agent receives a big punishment of $-\lambda$ whenever its action is not a member of $A_{\text{PSA}}$  which is also not necessarily suggested in \cite{alshiekh2017safe} and not used in \cite{Mirchevska_lane_change} for safety interference. 

One challenge here is choosing the safety interference punishment $\lambda$.
Lazarus et. al. in \cite{lazarus_runtime_safe_rl} applied similar rewarding scheme for an autopilot system of the aircraft which punished the RL agent whenever the emergency controller had to be deployed. 
They showed choosing higher values for $\lambda$ will encourage more conservative behavior, whereas faster policies with more interruptions can be expected for lower values of $\lambda$.

One may try to find the best value for $\lambda$, however, the learned policy generates proper behavior only for  environments that have similar state transition probabilities to the training environment $\tm$. 
Moreover, due to the safety verification punishments which consider worst-case assumptions, RL either learns a super conservative policy or a fast policy with too many safety interventions.
In our experiments, we learned a balanced policy by DQN with $\lambda=1$ as a normal RL baseline.

\subsection{Risk-aware Policies with Distributional Reinforcement Learning}\label{sec:iqn_cvar}
The main problem with applying normal RL in environments under uncertainty is its risk-neutral characteristic that can not distinguish the amount of variance in actions' return and only considers their expected values.
Therefore, one risky action with negative tail in its return but higher total expected value is always preferred to a safer (lower variance) action.
For that, Tang et al. modeled the return for each action as a normal distribution and optimized an RL policy by learning the return distribution parameters ($\mu$,$\sigma$) \cite{tang2020worst}.
However, we believe that the return is a multimodal distribution and therefore we applied Implicit Quantile Networks (IQN) implementation \cite{dabney2018implicit} which approximates the quantile function that implicitly defines the return distribution.  
Moreover, utilizing the return distribution also allows to expand risk-neutral policies to risk-sensitive policies by applying distortion risk measures like the Conditional Value-at-Risk (CVaR) \cite{cvar2000Rockafeller}:
\begin{equation}
\text{CVaR}_{\alpha} = \mathbb{E} \left[ Z^{\pi} | Z^{\pi} \leq F^{-1}_{Z^{\pi}}(\alpha) \right],
\end{equation}
where $Z^{\pi}$ is the sum of discounted rewards under policy $\pi$ and $F^{-1}_{Z^{\pi}}(\alpha)$ is its quantile function at $\alpha\in\left[0, 1\right]$.

Therefore, we train an IQN agent with same reward function as DQN (Equation \ref{eq:reward} with $\lambda=1$ ) and tune the the risk-sensitivity of the policy using $\alpha$ at execution time without requiring to train multiple policies for different risk levels.
In order to leverage the whole solution space over different risk-sensitive policies, $\alpha$ is uniformly sampled with $\alpha \sim U(0,1)$ at the beginning of each episode during training and applied to the IQN results (Figure \ref{fig:architectures}).

\section{Results and Evaluations}
\label{sec:results_and_evaluation}
\begin{figure*}[t]
	\minipage{0.32\textwidth}
	\def\svgwidth{\linewidth}
	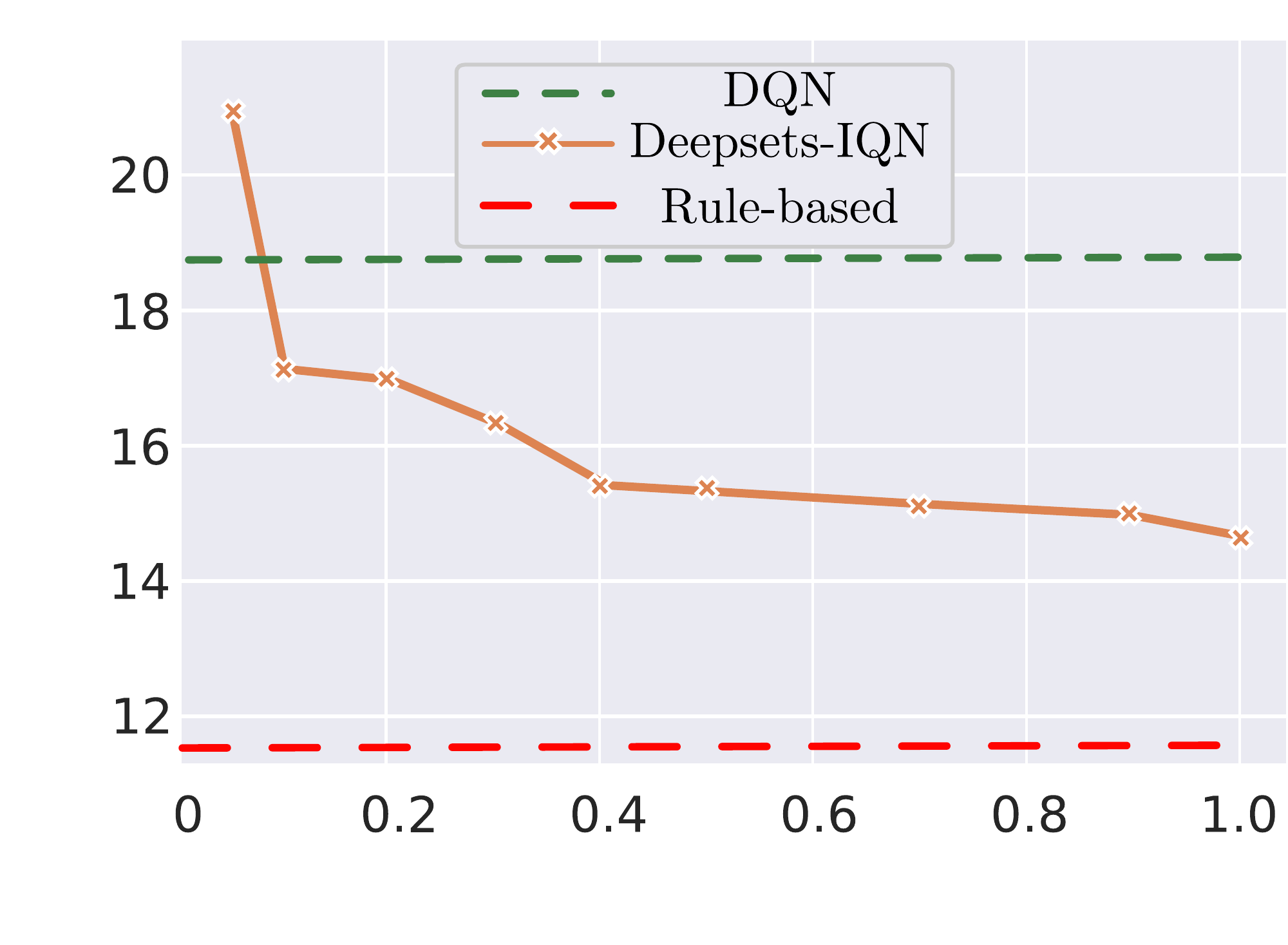
	\endminipage\hfill
	\minipage{0.32\textwidth}
	\def\svgwidth{\linewidth}
	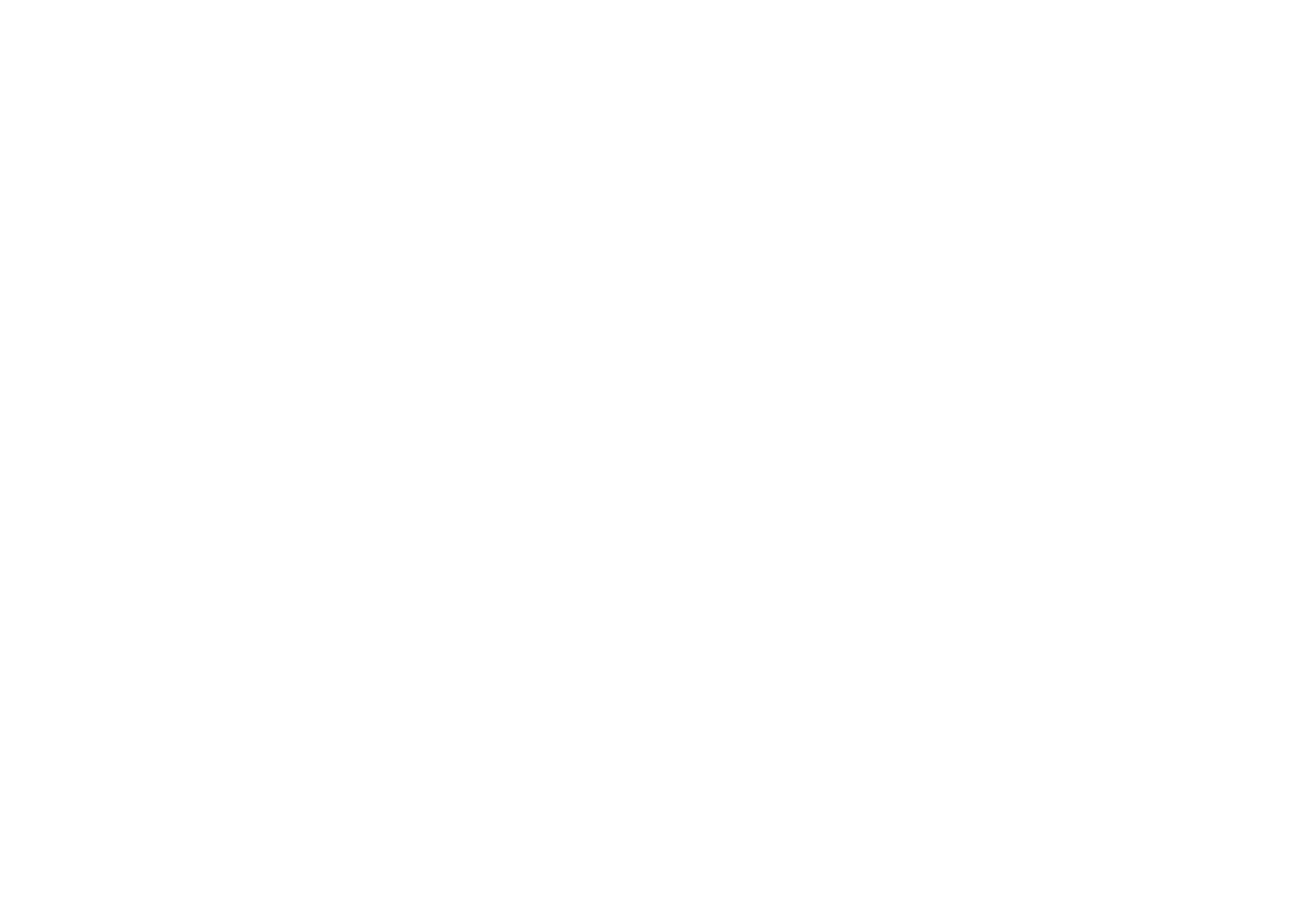
	\endminipage\hfill
	\minipage{0.32\textwidth}
	\def\svgwidth{\linewidth}
	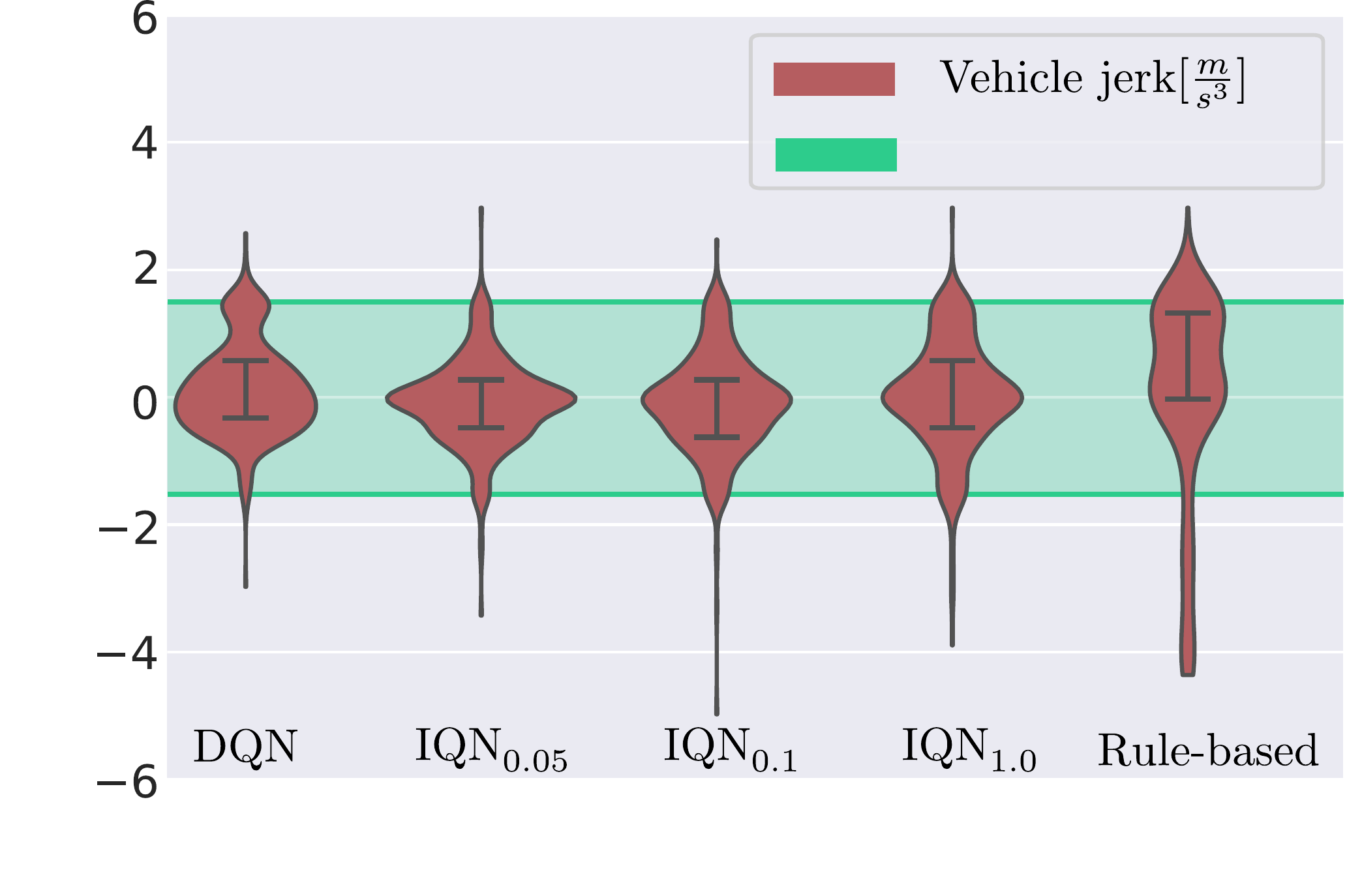
	\endminipage
	\caption{
		Comparing performance of each policy on benchmark scenarios.
		\textbf{Left:} Average crossing time for each policy. IQN becomes faster and less conservative by higher $\alpha$. Rule-based is the fastest and DQN is the slowest policy. Note that all policies are completely safe without any collision thanks to the safety layer.
		\textbf{Middle:} Average of vehicle absolute jerk and its confidence interval for each policy. By increasing $\alpha$, IQN becomes less conservative (higher jerk). 
		\textbf{Right:} Distribution of vehicle jerk while driving by each policy. Green interval shows RL jerk limits. Jerk values out of this interval are due to emergency interference and indicate uncomfortable driving.
	}\label{fig:all_eval}
\end{figure*}

\begin{figure}[t]\centering
	\fontsize{8pt}{11pt}\selectfont
	\def\svgwidth{0.99\linewidth}
	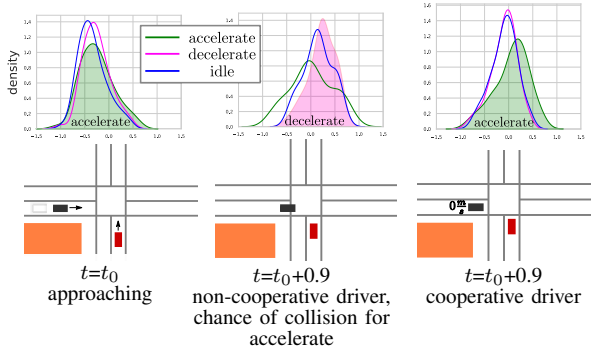
	\caption{\textbf{Left:} Learned return distributions for two scenarios with similar initial situation. Ego vehicle (red) follows the action with the highest CVaR that is shown with filled distribution. 
	Gray ghost vehicle indicates maximum visible distance limited due to orange obstacles.
	\textbf{Middle:} Black vehicle is non-cooperative causing more negative return chance for the \textit{accelerate} action.
	\textbf{Right:} Black vehicle is cooperative causing higher CVaR for \textit{accelerate} action. }\label{fig:scenarios_nice}
\end{figure}

\subsection{Simulation and Policy Training}
We train and evaluate different RL policies for automated merging and crossing at occluded intersections using our high level simulator which can simulate different randomized scenarios.
Figure \ref{fig:scenarios_nice} shows an example scenario generated by our simulator.
The location and size of obstacles (orange boxes) are generated randomly for each training episode causing some vehicles invisible for the ego vehicle.
In order to generate realistic scenarios, every vehicle in the simulator (except the ego vehicle) has a random desired velocity selected by normal distribution with a randomly selected $\mu$=\{6, 9, 12\} and $\sigma$=\{2, 4, 6\}.
Each vehicle drives with the Intelligent Driver Model \cite{treiber2000IDM} controller with maximum acceleration  $1\mpsq$, minimum deceleration $10\mpsq$ and comfortable deceleration $1.6\mpsq$ and keeps safety distance $2 \text{m}$ and time headway $1.6$s with front vehicles.
However, the controller does not keep distance to the ego vehicle (when it drives into the intersection), thus a collision between a vehicle and the ego vehicle is possible.

At every step a new vehicle by probability of $p_{\text{new}}$=\{0.1, 0.4 or 0.7\} is generated in the simulator.
Vehicles are with probability $p_{\text{coop}}$=\{0.1, 0.4 or 0.7\} cooperative and yield to the ego vehicle by setting their desired velocity to zero when the ego vehicle is close to the intersection which helps to simulate uncertainty about the intention of other drivers.
Moreover, distance and velocity of vehicles are perturbed to simulate perception noise for the policy according to the truncated Gaussian distributions 
$\mathcal{N}_3^t(0,\,\sigma_d^{2})$ 
and 
$\mathcal{N}_3^t(0,\,\sigma_v^{2})$,
respectively, where $\sigma_d=1\text{m}$ and $\sigma_v=2\mps$ were fixed during training.
$\sigma_d$ and  $\sigma_v$ are linearly scaled according to the distance between the ego and other vehicle.
For the ego vehicle, the RL policy generates a discrete jerk action from $\A=\{-1.5\mpsc, 0.0\mpsc, 1.5\mpsc \}$ every $0.3$ second.
RL policies are trained with more than 1000 intersection crossing and merging scenarios providing more than $3\times 10^5$ training steps in total.
Distributional RL policies have been trained with uniformly selected $\alpha \sim U([0,1])$ for CVaR calculation of the return distributions.


\subsection{Evaluations}
After training the RL agents, we evaluate their efficiency using 30 benchmark scenarios with random uncertainty configurations $\sigma_d\in\{0,1m,2m\}$, $\sigma_v=2\sigma_d$ and $p_\text{coop}$=\{0.1, 0.4 or 0.7\}. 
Note that, no collision with other vehicles happens during evaluations thanks to the safety verification layer.
In case of an emergency situation, i.e. $a_\text{RL} \notin A_{\text{PSA}}$, an emergency maneuver from the safety layer with maximum emergency jerk limit of $5\mpsc$ is sent to the controller instead of the unsafe RL action.
In order to compare the effect of interference applied for each policy, we define a metric for measuring the amount of applied emergency interference:
\begin{equation}
J_\text{Interference} = \frac{\sum a_\text{emg}^2}{N},
\end{equation}
where $a_\text{emg}$ is the emergency jerk command applied to replace the unsafe action and $N$ is the total number of evaluation episodes.

\subsubsection{Drivers Intention Prediction}
Since RL policies receive history of last 5 observations as their input state, they can predict the intention of other drivers and generate optimal actions based on that. 
Figure \ref{fig:scenarios_nice} shows examples of the learned return distributions by IQN agent for two similar scenarios with the only difference of having cooperative (yielding to the ego vehicle) and non-cooperative drivers.
As it is visible, when the other vehicle is cooperative and reduces its velocity, the IQN policy generates higher return for $a=1.5\mpsc$ action allowing the ego vehicle to enter into the intersection.

\begin{table*}[t]\centering
	\caption{Evaluation of the IQN policies with best $\alpha$ for different metrics and their comparison with baselines for different environment configurations.}
	\begin{tabular}{l|l|ll|ll|ll}
		\hline
		\multicolumn{2}{l|}{Environment Configuration}        &
		\multicolumn{2}{l|}{$\sigma_d = 0$ $p_\text{coop} = 0.3$} &
		\multicolumn{2}{l|}{$\sigma_d=2$ $p_\text{coop} = 0.3$ } &
		\multicolumn{2}{l}{$\sigma_d=2$ $p_\text{coop} = 0.7$} \\ \Xhline{3\arrayrulewidth}
		Policy                       & Metric for optimal $\alpha$    & Time          & $J_\text{Interference}$       & Time       &  $J_\text{Interference}$         & Time         &  $J_\text{Interference}$        \\ \Xhline{3\arrayrulewidth}
		\multirow{2}{*}{Deepset-IQN} 
		& Speed   &   9.80   &   25.03          & 8.80    & 43.09   &  9.94       &    33.25       \\
		& Comfort &   11.73  &   \textbf{10.00} & 11.08   & 25.50   &  12.16      &   \textbf{18.83 }      \\ 
		\hline
		DQN & -----&  20.00           &  10.79  & 21.40    & \textbf{14.17}   &     21.25      &     20.10  \\ 
		\hline
		Rule-based & -----&  \textbf{7.69}   &  83.59  & \textbf{7.75}     & 92.29   &  \textbf{7.74} & 103.05  \\ 
		\hline
	\end{tabular}\label{table:time_cost}
\end{table*}

\subsubsection{Comfort and Risk Sensitivity}
Figure \ref{fig:all_eval} compares the utility and conservativity of each RL policy compared with the safe rule-based policy.
By increasing the $\alpha$ percentile, IQN policy becomes faster but less conservative in average.
For lower $\alpha$ percentile, it has less absolute jerk (in the middle image) and lower jerks outside the RL jerk limit (in the right image) which indicates more comfortable maneuvers.
DQN, shows the most conservative behavior with the highest crossing time.
On the other hand, the rule-based policy has the highest amount of emergency interference as a risk neutral policy resulting in wider jerk distributions and showing uncomfortable maneuvers.
Therefore, we can conclude that IQN can learn a \textit{family} of adaptive policies which are not as highly conservative as DQN and also not as risk-neutral as the rule-based policy.

\begin{figure}[t]\centering
	\fontsize{8pt}{11pt}\selectfont
	\def\svgwidth{0.83\linewidth}
	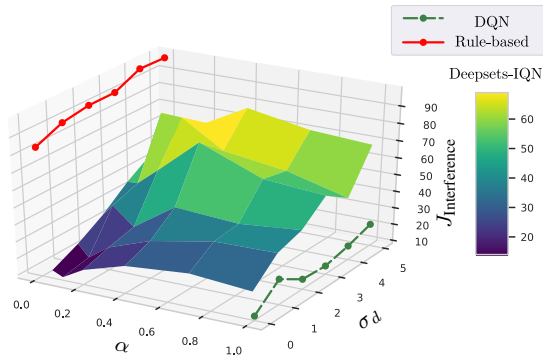
	\caption{Safety interference cost applied to the IQN policy for different environment noise levels and its comparison with baselines.}\label{fig:alpha_noise_cost}
\end{figure}

Same results can be concluded from Table \ref{table:time_cost} where we recorded the average crossing time and $J_\text{Interference}$ for all policies at different environment configurations.
Here we only consider two IQN sub-policies, the ones with the fastest behavior and with the lowest interference cost (most comfortable).
We evaluated the policy in 10 uniform samples of $\alpha$ and selected the best policy according to each metric.
Note that finding the best $\alpha$ value could be done in a more automated approach but here we only wanted to compare the best behavior of the Deepset-IQN agent with other baseline agents and examine how much other metrics are sacrificed when the policy performs well in specific metric.
In all three configurations considered in Table \ref{table:time_cost}, the rule-based policy is the fastest one, however, it is also the worst policy in terms of comfort.
The DQN, on the other hand, always has low interference cost, but drives more than two times slower than the others.
The IQN trades of between average crossing time and interference cost by showing relatively fast behavior which requires 3.6 to 8 times lower interference than the rule-based policy depending on the environment configuration.

\subsubsection{Evaluation under Higher Perception Uncertainties}
In addition to the initial noise levels, we studied the sensitivity of each policy to higher amounts of noise in the environment.
For that, we evaluated the RL and rule-based policies for 5 different noise levels $\sigma_d\in\{0,1,2,3,4,5\}$ (in meters) and $\sigma_v=2\sigma_d$ (in $\mps$).
The results are depicted in Figure \ref{fig:alpha_noise_cost}.
As it is visible, the IQN policy has lower interference cost for lower $\alpha$ and also lower noise levels.
For higher noise levels, it requires more interference which can be reduced by decreasing CVaR $\alpha$.
The rule-based policy always has high cost and DQN has low cost independent of the amount of noise existed in the environment.
Here DQN seems to be a comfortable policy, however it is the worst policy in case of utility as discussed before.

\section{Conclusions and Future Work}
\label{sec:conclusion}

We proposed a proactive safety verification approach to validate the safety of actions with the goal of preventing unsafe situations in critical applications like automated driving.
In order to minimize uncomfortable safety interference, we used reinforcement learning to learn policies that are punished for resulting in states where an emergency maneuver is necessary.
We showed how IQN agent can mitigate the conservative behavior existing in DQN policies by learning return distributions which provide policies that can be tuned adaptively after training in order to become more comfortable or less conservative.
According to our experiments, the learned distributional policy requires less safety interference in noisy environments comparing to a rule-based safe policy even when the noise level is higher than the training configurations.

For future works, one can propose a better conservativity tuning approach in order to learn policies that automatically tune their risk sensitivity ($\alpha$ for the IQN policy) at runtime based on the observed uncertainty or the required risk sensitivity in the environment.
\section{Acknowledgment}
This research is accomplished within the project ``UNICARagil'' (FKZ 6EMO0287).
We acknowledge the financial support for the project by the Federal Ministry of Education and Research of Germany (BMBF).

\printbibliography

\end{document}